\documentclass[10pt]{article}

\usepackage{PRIMEarxiv}

\usepackage[utf8]{inputenc} 
\usepackage[T1]{fontenc}    
\usepackage{hyperref}       
\usepackage{url}            
\usepackage{booktabs}       
\usepackage{amsfonts}       
\usepackage{nicefrac}       
\usepackage{enumitem}
\usepackage{lipsum}
\usepackage{multirow}
\usepackage[linesnumbered,ruled,vlined]{algorithm2e}
\SetKwInput{Input}{Input}   
\SetKwInput{Output}{Output} 
\usepackage{amsmath, amssymb, mathrsfs, bbm}
\usepackage{tabularx}
\usepackage{setspace}
\usepackage{subcaption}
\usepackage{xcolor}         
\usepackage{fancyhdr}       
\usepackage{graphicx}       
\usepackage[backend=biber,style=ieee,sorting=none,natbib=true]{biblatex}
\usepackage{authblk}

\graphicspath{{media/}}     
\usepackage{bm}

\title{\textbf{From Orthomosaics to Raw UAV Imagery: Enhancing Palm Detection and Crown-Center Localization}}
\author[1,*]{Rongkun Zhu}
\author[2,*]{Kangning Cui}
\author[3]{Wei Tang}
\author[4]{Rui-Feng Wang}
\author[2]{Sarra Alqahtani}
\author[5]{David Lutz}
\author[2]{Fan Yang}
\author[6]{Paul Fine}
\author[7]{Jordan Karubian}
\author[2]{Robert Plemmons}
\author[8]{Jean\mbox{-}Michel Morel}
\author[2]{Victor Pauca}
\author[2]{Miles Silman}
\affil[1]{ Departments of Computer Science, Hong Kong Baptist University}
\affil[2]{ Departments of Computer Science \& Biology, Wake Forest University}
\affil[3]{ Department of Mathematics, City University of Hong Kong}
\affil[4]{ Department of Crop and Soil Sciences, University of Georgia}
\affil[5]{ School of Arts \& Sciences, Colby-Sawyer College}
\affil[6]{ Department of Integrative Biology, University of California, Berkeley}
\affil[7]{ Department of Ecology and Evolutionary Biology, Tulane University}
\affil[8]{ School of Data Science, Lingnan University}

\begin{document}
\maketitle

\let\thefootnote\relax
\noindent\footnotetext{$*$ Rongkun Zhu and Kangning Cui contributed equally to this work. Corresponding author: cuij@wfu.edu}

\begin{abstract}
Accurate mapping of individual trees is essential for ecological monitoring and forest management. Orthomosaic imagery from unmanned aerial vehicles (UAVs) is widely used, but stitching artifacts and heavy preprocessing limit its suitability for field deployment. This study explores the use of raw UAV imagery for palm detection and crown-center localization in tropical forests. Two research questions are addressed: (1) how detection performance varies across orthomosaic and raw imagery, including within-domain and cross-domain transfer, and (2) to what extent crown-center annotations improve localization accuracy beyond bounding-box centroids. Using state-of-the-art detectors and keypoint models, we show that raw imagery yields superior performance in deployment-relevant scenarios, while orthomosaics retain value for robust cross-domain generalization. Incorporating crown-center annotations in training further improves localization and provides precise tree positions for downstream ecological analyses. These findings offer practical guidance for UAV-based biodiversity and conservation monitoring.

\end{abstract}


\section{Introduction}
\label{sec:intro}

Palms (\textit{Arecaceae}) are ecologically and economically key components of tropical ecosystems. They play a vital role in shaping forest structure, supporting biodiversity by providing resources for wildlife, and sustaining local livelihoods~\cite{eiserhardt2011geographical,cui2025detection}. Their distinctive crowns and high visibility make them well-suited for automated mapping using unmanned aerial vehicle (UAV) imagery, which can aid in biodiversity monitoring and sustainable land management~\cite{sutherland2013identification, cui2025efficient, di2025toward}.

Previous works to localize palms in landscape scales have mainly relied on orthomosaic imagery, which is generated by stitching hundreds of aerial photographs together~\cite{gibril2021deep, ferreira2020individual}. While orthomosaics offer spatially consistent views, they often suffer from stitching artifacts, reduced resolution, and need extensive pre- and post-processing steps, limiting their applicability for on-field and edge deployments~\cite{zhang2023aerial, shao2025flashsvd, jiang2025sada}. These limitations are particularly pronounced in dense forests, where occlusions, variable illuminations, and irregular spatial distributions pose significant challenges for object detection models~\cite{tagle2019identifying, cui2024palmprobnet, zhang2025center, wang2026cott}. In such settings, overlapping canopies often cause systematic offsets between predicted box centers and the crown centers, thereby reducing localization accuracy and limiting the reliability of downstream ecological analyses.

In this work, we move beyond \textit{orthomosaics} by applying \textit{raw UAV imagery} to improve both visual fidelity and computational efficiency. Raw images preserve full-resolution detail without stitching artifacts, minimize pre-processing, and support flexible deployment in real-time or low-resource scenarios. To evaluate the trade-offs, we compare detection performance across raw and orthomosaic imagery under both within- and cross-domain settings. We also address systematic offsets between bounding-box centroids and crown centers in dense forests by introducing explicit crown-center annotations, which provide ecologically meaningful references for spatial analysis. Our contributions are threefold: (1) a new dataset and annotation protocol in raw UAV imagery that includes both bounding boxes and crown centers; (2) a cross-domain study that analyzes trade-offs under distribution and image-domain shifts; and (3) an extension of our previous PRISM pipeline~\cite{cui2025detection} with improved crown-center localization.

\section{Related Work}
\label{sec:related}

\subsection{Palm Detection and Localization}

Research on palm detection has evolved from traditional machine learning to deep learning approaches, as also seen more broadly in remote sensing~\cite{yang2026superpixel, cui2024superpixel, polk2023unsupervised, camalan2022change}. The early methods were often based on rules or relied on sliding windows with convolutional networks, and were developed primarily for structured plantations such as oil palms and date palms~\cite{li2016deep, gibril2021deep}. Although these approaches achieved reasonable performance in uniform settings, they often required extensive post-processing and struggled to generalize to natural forests. In parallel, object detection itself has advanced rapidly, moving from early region-based methods to efficient single-shot detectors and, more recently, transformer-based architectures, greatly expanding the toolkit for ecological applications~\cite{zou2023object}. Although recent progress in UAVs and high–resolution imagery has lowered the barrier to analyzing naturally occurring palms at landscape scales, the task remains far from straightforward, as heterogeneous canopies and complex backgrounds continue to pose challenges~\cite{cui2025detection}.

Motivated by these advances, studies have shifted from structured plantations to natural tropical forests. Palm crowns are identified and delineated from UAV imagery in~\cite{tagle2019identifying} using an object-based workflow that integrates color and textural features with GIS tools. Individual tree detection and species classification of Amazonian palms are performed in~\cite{ferreira2020individual} by applying a fully convolutional network to UAV images with morphological refinements to separate overlapping crowns. At the regional scale, canopy palm mapping is carried out in~\cite{wagner2020regional} using a U-Net model with high-resolution multispectral satellite imagery. Our previous work follows this line of research: PRISM introduces a modular pipeline that integrates detection, segmentation, and calibration, and was evaluated on a multi-site UAV orthomosaic dataset from 21 Ecuadorian locations~\cite{cui2025detection}. Extending this trajectory, we further coupled PRISM with a Poisson–Gaussian reproduction model to simulate spatial distributions in wild forests~\cite{cui2025efficient}.

\subsection{Keypoint Detection}

Keypoint detection provides fine-grained spatial localization by predicting semantically meaningful points, such as object centers or structural landmarks~\cite{fan2022deep}. Unlike bounding boxes, which provide coarse spatial extent, keypoints enable more precise localization and are particularly useful in dense or irregular scenes where objects overlap~\cite{maji2022yolo}. Recent studies demonstrate their utility in tasks such as cross-frame tracking~\cite{feng2021cross} and precision agriculture applications~\cite{meng2025yolov10, zhao2024automatic}. Broader surveys highlight their strengths in pose estimation and tracking under occlusion and cluttered conditions~\cite{fan2022deep}. 

However, keypoint methods remain underexplored in ecological monitoring, where bounding boxes are the dominant annotation form. Crown centers provide more ecologically meaningful references than box centroids in dense forests, which directly link to tree locations needed for spatial distribution modeling and downstream ecological analyses. This makes crown-center detection a natural extension of keypoint frameworks to ecological applications.

\section{Data and Method}
\label{sec:data_method}
\subsection{Study Site and Data Collection}

\begin{figure}[t]
    \centering
    \includegraphics[width=0.7\linewidth]{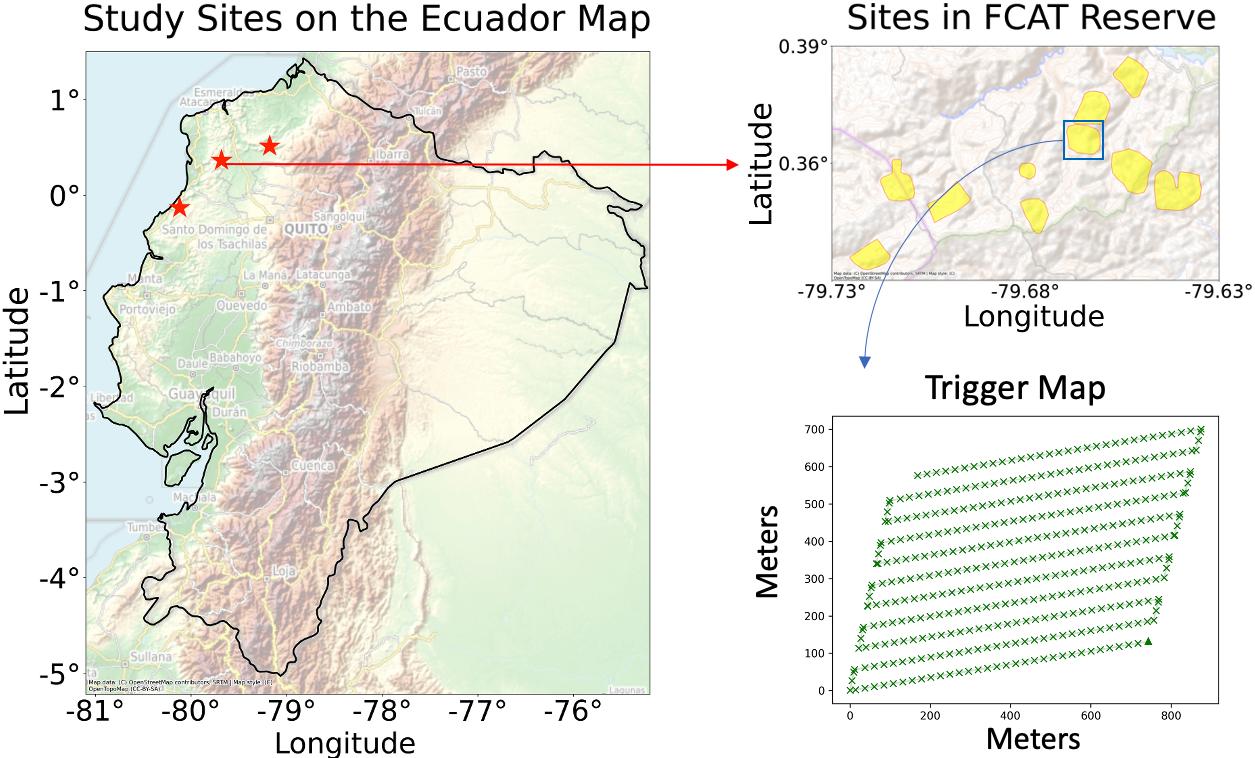}
    \caption{Study site and UAV survey at FCAT. Left: location in Ecuador. Top-right: survey block for raw image annotations. Bottom-right: trigger map of image captures at 90 m altitude.}
    \label{fig:map_paths}
\end{figure}

This study was conducted at the \textit{Fundación para la Conservación de los Andes Tropicales Reserve and adjacent Reserva Ecológica Mache-Chindul park} (FCAT) in western Ecuador (00$^\circ$23'28'' N, 79$^\circ$41'05'' W). FCAT lies within the Chocó region and represents a highly diverse humid tropical forest at $\sim$500 m elevation. The site receives approximately 3,000 mm of rain annually, with frequent fog during the drier months. Several palm species with exposed canopy crowns are abundant in FCAT, with dominant species \textit{Iriartea deltoidea} and \textit{Socratea exorrhiza}~\cite{browne2016diversity,lueder2022functional}. UAV imagery was collected as part of the 2022 survey campaign described in~\cite{cui2025detection,cui2025efficient}, using a DJI Phantom 4 RTK drone equipped with a 1-inch CMOS sensor. The flights were carried out at an altitude of 90 meters above ground level with automated mission planning to ensure consistent coverage and overlap (See Figure~\ref{fig:map_paths} for the study site and the flight paths). The top-right inset highlights the survey block from which the raw UAV images were newly annotated for this study, while the bottom-right trigger map shows the sequence of image captures during the mission.  

\subsection{Datasets and Annotation Protocol}

In this study, two UAV-based palm tree datasets collected at three FCAT sites were used. Although these sites share similar ecological conditions, the datasets differ in the image domain and annotation richness. \textit{Orthomosaic Patches}\footnote{Available at: \href{https://github.com/Zippppo/PRISM}{github.com/Zippppo/PRISM}}~\cite{cui2025detection} consists of 1,500 images (800$\times$800 pixels) cropped from two large-scale orthomosaics covering two FCAT sites, with 8,842 annotated palm bounding boxes ($\sim$5.89 per image). Since annotations are limited to bounding boxes, this dataset is relatively simple and is suitable for detection tasks. 

In contrast, \textit{Raw Patches}\footnote{Datasets available at: \href{https://zenodo.org/records/17094346}{zenodo.org/records/17094346}} contain 880 images (912$\times$912 pixels) with 5,850 bounding boxes and 5,430 crown centers ($\sim$6.65 per image). Newly annotated for this study, the dataset follows a protocol in which the bounding boxes delineate the palm crowns and each palm is assigned a crown center when visible. Crown centers indicate the geometric center of the visible crown, which may deviate from the bounding-box center in cases of occlusion or boundary truncation. These additional annotations enable for a finer-grained basis for evaluating localization accuracy.

\subsection{Research Questions and Method}

This study addresses two main research questions (RQs): 

\noindent\textit{RQ1: How does model performance vary when training and testing on orthomosaic versus raw UAV imagery, including within-domain and cross-domain transfer?}  
This comparison is critical for ecological monitoring workflows, where orthomosaics provide standardized mosaicked views while raw images retain higher geometric fidelity. We compare within-domain and cross-domain scenarios to evaluate domain transferability and assess whether models trained on one image domain can be directly applied to the other.  

\noindent\textit{RQ2: To what extent do crown-center annotations improve palm localization beyond bounding-box centroids?}  
While bounding boxes capture coarse crown extents, crown centers provide a more precise geographic reference for ecological applications. We assess whether explicitly annotated crown centers reduce localization error and quantify their improvement under varying conditions such as occlusion and boundary truncation. This serves as a proxy for potential benefits in downstream tasks that need accurate tree positions. 

To investigate these RQs, we employ state-of-the-art detection frameworks under consistent training protocols. YOLOv8–12 and RT-DETRv1–v2 are used for crown detection, while YOLOv8, YOLOv11, and YOLOv12 in pose mode are used for crown-center localization~\cite{yolo11, lv2023detrs, lv2024rtdetrv2}.

\begin{table*}[t]
\centering
\scriptsize
\setlength{\tabcolsep}{3pt}
\caption{Detection performance of RT-DETR and YOLO variants evaluated with COCO-style metrics (mAP and mAR at IoU thresholds 0.5:0.95, 0.5, and 0.75). Results are reported under two training regimes (Train=Raw and Train=Orthomosaic) and two test regimes (Test=Raw and Test=Orthomosaic), covering both within-domain and cross-domain settings. 
Bold values indicate the best score per test condition, and underlined values indicate the second best.}
\resizebox{\textwidth}{!}{
\begin{tabular}{l c | cccccc | cccccc}
\toprule
\multirow{2}{*}{\textbf{Model}} & \multirow{2}{*}{\textbf{Test}}
& \multicolumn{6}{c|}{\textbf{Train = Raw}}
& \multicolumn{6}{c}{\textbf{Train = Orthomosaic}} \\
\cmidrule(lr){3-8}\cmidrule(lr){9-14}
& & \textbf{mAP} & \textbf{mAP@0.5} & \textbf{mAP@0.75} & \textbf{mAR} & \textbf{mAR@0.5} & \textbf{mAR@0.75}
  & \textbf{mAP} & \textbf{mAP@0.5} & \textbf{mAP@0.75} & \textbf{mAR} & \textbf{mAR@0.5} & \textbf{mAR@0.75} \\
\midrule
\multirow{2}{*}{RT-DETRv1}
& Raw & 0.594 & 0.887 & 0.656 & 0.684 & 0.963 & 0.746
      & 0.534 & \underline{0.864} & 0.583 & 0.632 & 0.942 & 0.698 \\
& Orthomosaic & \underline{0.389} & \textbf{0.662} & 0.392 & 0.535 & \textbf{0.872} & 0.549
      & 0.571 & 0.859 & 0.609 & 0.674 & 0.969 & 0.723 \\
\midrule
\multirow{2}{*}{RT-DETRv2}
& Raw & 0.607 & 0.916 & 0.684 & 0.699 & 0.972 & \underline{0.797}
      & \underline{0.546} & 0.860 & \underline{0.621} & \textbf{0.655} & \textbf{0.953} & \textbf{0.745} \\
& Orthomosaic & 0.363 & 0.631 & 0.361 & 0.526 & 0.833 & 0.544
      & 0.613 & 0.906 & 0.668 & \textbf{0.713} & \textbf{0.975} & \textbf{0.790} \\
\midrule
\multirow{2}{*}{YOLOv8}
& Raw & 0.615 & 0.913 & 0.691 & 0.702 & 0.982 & 0.779
      & 0.530 & 0.837 & 0.587 & 0.610 & 0.900 & 0.675 \\
& Orthomosaic & \textbf{0.398} & \underline{0.658} & \textbf{0.407} & \textbf{0.547} & 0.845 & \textbf{0.576}
      & 0.608 & 0.897 & 0.662 & 0.679 & 0.946 & 0.738 \\
\midrule
\multirow{2}{*}{YOLOv9}
& Raw & 0.615 & 0.910 & 0.699 & 0.695 & 0.969 & 0.787
      & \textbf{0.550} & 0.853 & \textbf{0.626} & 0.637 & 0.924 & \underline{0.717} \\
& Orthomosaic & 0.374 & 0.636 & 0.384 & 0.497 & 0.795 & 0.516
      & 0.618 & \underline{0.908} & \underline{0.678} & 0.686 & 0.955 & 0.750 \\
\midrule
\multirow{2}{*}{YOLOv10}
& Raw & 0.602 & 0.894 & 0.670 & 0.680 & 0.955 & 0.764
      & 0.511 & 0.824 & 0.548 & 0.599 & 0.884 & 0.646 \\
& Orthomosaic & 0.372 & 0.619 & 0.378 & 0.483 & 0.747 & 0.509
      & 0.607 & 0.895 & 0.670 & 0.676 & 0.939 & 0.744 \\
\midrule
\multirow{2}{*}{YOLO11}
& Raw & \textbf{0.627} & \textbf{0.928} & \textbf{0.713} & \underline{0.707} & \textbf{0.987} & \textbf{0.805}
      & 0.526 & 0.849 & 0.572 & 0.617 & 0.911 & 0.675 \\
& Orthomosaic & 0.379 & 0.649 & 0.385 & 0.529 & 0.839 & \underline{0.565}
      & \textbf{0.625} & \textbf{0.908} & \textbf{0.683} & 0.696 & 0.961 & 0.766 \\
\midrule
\multirow{2}{*}{YOLO12}
& Raw & \underline{0.623} & \underline{0.924} & \underline{0.704} & \textbf{0.711} & \underline{0.985} & 0.795
      & 0.545 & \textbf{0.873} & 0.592 & \underline{0.644} & \underline{0.952} & 0.714 \\
& Orthomosaic & 0.380 & 0.636 & \underline{0.393} & \underline{0.539} & \underline{0.860} & 0.564
      & \underline{0.623} & 0.905 & 0.677 & \underline{0.706} & \underline{0.971} & \underline{0.767} \\
\bottomrule
\end{tabular}}
\label{tab:detection_results_mAR}
\end{table*}

\section{Experiments}

\subsection{Experiment Settings and Metrics}

Both datasets were divided into training, validation, and testing sets using an 80/10/10 ratio. For the \textit{Orthomosaic Patches}, the split is 1,200/150/150 images, while for the \textit{Raw Patches} the split is 704/88/88 images. All models were trained for 300 epochs on a single NVIDIA L40S GPU.

For palm detection, we adopted COCO-style metrics, reporting mean Average Precision (mAP) and mean Average Recall (mAR) with up to 100 detections per image. Metrics were averaged across IoU thresholds from 0.50 to 0.95, with additional reports at 0.50 and 0.75. Detection experiments include YOLOv8–12 and RT-DETRv1–v2, using the largest available variants. For crown-center localization, we followed the Ultralytics YOLO Pose evaluation protocol. Metrics include keypoint mAP (mAP\textsubscript{pose}) over Object Keypoint Similarity (OKS) thresholds (0.50–0.95),  with additional results at 0.50 and 0.75, together with precision, recall, and F1 score of predicted centers. The experiments used YOLOv8, YOLO11, and YOLO12 in pose mode, chosen for their strong detection performance and built-in support for keypoint estimation.

\begin{table}[t]
\centering
\scriptsize
\setlength{\tabcolsep}{4pt}
\caption{Crown-center localization results with YOLO pose models. Each model is evaluated using box centers and predicted crown centers. Metrics include mAP over OKS thresholds (50–95, 50, 75), precision, recall, and F1-score.}
\resizebox{0.6\linewidth}{!}{
\begin{tabular}{lccccccc}
\toprule
\textbf{Model} & \textbf{Method} & \textbf{mAP} & \textbf{mAP@50} & \textbf{mAP@75} & \textbf{Precision} & \textbf{Recall} & \textbf{F1} \\
\midrule
\multirow{2}{*}{YOLOv8-Pose}
& Box Center   & 0.316 & 0.520 & 0.327 & 0.625 & 0.489 & 0.549 \\
& Crown Center   & 0.767 & \textbf{0.914} & \textbf{0.839} & \textbf{0.873} & \textbf{0.848} & \textbf{0.860} \\
\midrule
\multirow{2}{*}{YOLO11-Pose}
& Box Center   & 0.340 & 0.566 & 0.332 & 0.719 & 0.518 & 0.602 \\
& Crown Center   & 0.748 & 0.869 & 0.803 & 0.805 & 0.800 & 0.803 \\
\midrule
\multirow{2}{*}{YOLO12-Pose} 
& Box Center   & 0.318 & 0.532 & 0.317 & 0.612 & 0.528 & 0.567 \\
& Crown Center   & \textbf{0.769} & 0.906 & 0.822 & 0.870 & 0.829 & 0.849 \\
\bottomrule
\end{tabular}}
\label{tab:keypoint_results}
\end{table}

\subsection{Result Analysis}

We first address \textit{RQ1}, which examines how model performance varies when training and testing on orthomosaic versus raw UAV imagery. Table~\ref{tab:detection_results_mAR} summarizes the results across within- and cross-domain settings. When the ultimate goal is on-field deployment with raw UAV imagery, models trained and tested on \textit{Raw Patches} achieve the best performance. For example, YOLOv11x reaches 0.627 mAP and 0.707 mAR, while YOLOv12x closely follows at 0.623 mAP and 0.711 mAR. These results exceed the best orthomosaic-trained models evaluated on raw images by more than 6\% in both mAP and mAR, confirming that raw imagery provides higher-quality training signals and better detection accuracy in deployment-relevant settings. Within the \textit{Orthomosaic Patches}, models also perform competitively despite noise and stitching artifacts. RT-DETRv2, YOLOv11x, and YOLOv12x all surpass 0.61 mAP and 0.69 mAR when trained and tested on orthomosaics. This highlights the utility of orthomosaic-based training for landscape-scale monitoring, where mosaicked imagery is commonly used.

Orthomosaic-trained models also transfer more effectively to raw images. YOLOv12x, for example, achieves 0.545 mAP on raw test data when trained on orthomosaics, compared to just 0.380 mAP in the reverse case. This asymmetry reflects the more diverse yet noisier orthomosaic annotations, which encourage better generalization to raw inputs. Conversely, raw-trained models degrade when applied to orthomosaics, in part due to their reliance on high-fidelity training signals, but also because the raw dataset is smaller in size, further limiting transferability. Since our focus is on raw imagery deployment, this limitation is less critical. Overall, these findings reveal a trade-off: orthomosaic training offers better cross-domain robustness, but raw training delivers the highest accuracy on raw imagery, which is essential for on-site and edge applications.

\begin{figure}[t]
    \centering
    \includegraphics[width=0.6\linewidth]{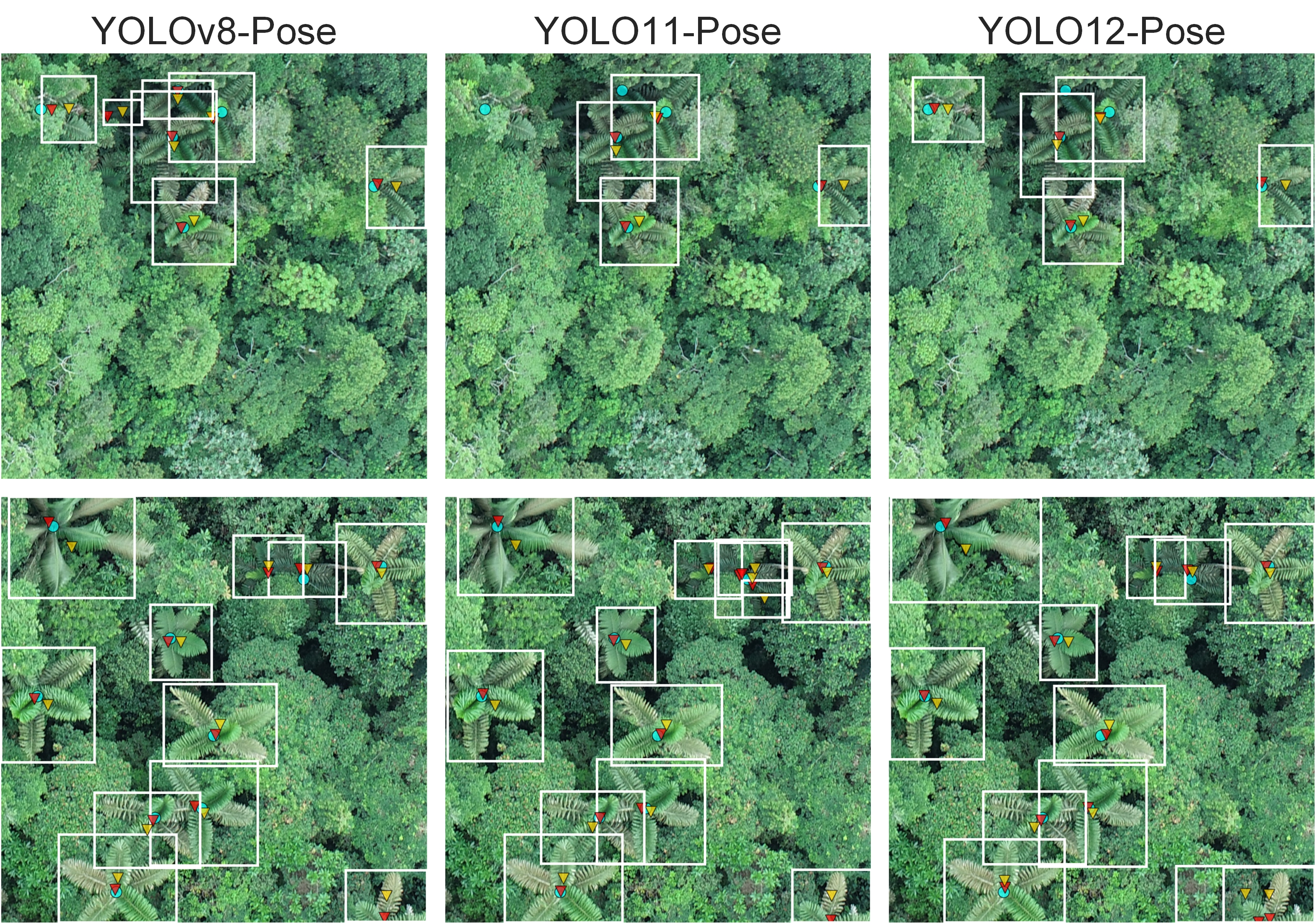}
    \caption{Comparison of crown-center predictions across YOLO pose models. Each panel shows ground-truth centers (blue dots), predicted crown centers (red triangles), and predicted bounding box centroids (yellow triangles).}
    \label{fig:center}
\end{figure}

We next address \textit{RQ2}, which evaluates whether crown-center annotations improve palm localization beyond bounding-box centroids. Table~\ref{tab:keypoint_results} shows that in all models, the predicted crown centers substantially outperform the box centroids, with the mAP improving from approximately 0.32-0.34 to over 0.75 and F1-scores increasing from 0.55–0.60 to above 0.80. These gains demonstrate that crown-center annotations reduce systematic offsets and yield more accurate localization. Figure~\ref{fig:center} further illustrates the qualitative differences. The predicted crown centers align more closely with ground-truth centers than the bounding-box centroids, particularly in dense or partially occluded canopies where bounding-box centroids are consistently displaced. Together, these results confirm that crown-center predictions provide a more reliable spatial reference for downstream ecological analyses.

\section{Conclusion and Future Work}

We presented a comparative study of palm detection and crown-center localization from UAV imagery, contrasting orthomosaic and raw domains. The results show that raw imagery yields higher accuracy for deployment, while orthomosaics provide robustness in cross-domain transfer. Crown-center annotations further reduce localization error and produce ecologically meaningful tree positions.

A natural extension is to combine mixed-domain training with inference on raw images. The detected crown centers can be georeferenced using the same transformations applied in orthomosaic generation, enabling spatial mapping without constructing full mosaics. This facilitates on-field and near-real-time monitoring, which traditional orthomosaic-based pipelines cannot achieve. Because crown-center detection fits seamlessly into the existing PRISM framework~\cite{cui2025detection}, it can directly enhance downstream spatial analyses. Future work will explore lightweight models for UAV on-board inference and extend crown-center outputs to palm spatial modeling.


\printbibliography 

\end{document}